\documentclass[conference]{IEEEtran}
\IEEEoverridecommandlockouts
\usepackage{cite}
\usepackage{todonotes}
\usepackage{amsmath,amssymb,amsfonts}
\usepackage{algorithmic}
\usepackage{graphicx}
\usepackage{url}
\usepackage{textcomp}
\usepackage{xcolor}
\usepackage{makecell}
\usepackage{tcolorbox}
\usepackage{threeparttable}
\usepackage{caption}
\usepackage{booktabs,siunitx,tabularx, stfloats}
\usepackage{array}
\usepackage{multirow}
\usepackage{adjustbox}
\usepackage{hyperref}
\usepackage{fancyhdr}
\usepackage{flushend}

\def\BibTeX{{\rm B\kern-.05em{\sc i\kern-.025em b}\kern-.08em
    T\kern-.1667em\lower.7ex\hbox{E}\kern-.125emX}}

\begin{document}

\title{Take It Easy: Label-Adaptive Self-Rationalization for Fact Verification and Explanation Generation
\thanks{This research was supported by São Paulo Research Foundation (FAPESP) under the grants Horus \#2023/12865-8 and \#2019/04053-8.}
}

\author{Jing Yang and Anderson Rocha\\
Artificial Intelligence Lab. (\url{Recod.ai}), Institute of Computing, University of Campinas, SP, Brazil \\
\{jing.yang.ic, anderson.rocha\}@unicamp.br\\}

\maketitle

\fancypagestyle{firstpage}{
    \renewcommand{\headrulewidth}{0pt}%
    \fancyhf{}
    \fancyfoot[L]{\footnotesize\textcopyright 2024 IEEE.  Personal use of this material is permitted. Permission from IEEE must be obtained for all other uses, in any current or future media, including reprinting/republishing this material for advertising or promotional purposes, creating new collective works, for resale or redistribution to servers or lists, or reuse of any copyrighted component of this work in other works.}
 }

\thispagestyle{firstpage}

\begin{abstract}
Fact verification is a crucial process in journalism for combating disinformation. Computational methods to aid journalists in the task often require adapting a model to specific domains and generating explanations. However, most automated fact-checking methods rely on three-class datasets, which do not accurately reflect real-world misinformation. Moreover, fact-checking explanations are often generated based on text summarization of evidence, failing to address the relationship between the claim and the evidence. To address these issues, we extend the self-rationalization method—typically used in natural language inference (NLI) tasks—to fact verification. Self-rationalization refers to a model's ability to generate explanations or justifications for its responses, which is essential for reliable fact-checking. We propose a label-adaptive learning approach: first, we fine-tune a model to learn veracity prediction with annotated labels (step-1 model). Then, we fine-tune the step-1 model again to learn self-rationalization, using the same data and additional annotated explanations. This approach allows the model to adapt to a new domain more effectively than fine-tuning end-to-end self-rationalization directly. Our results show that our label-adaptive approach improves veracity prediction by more than ten percentage points (Macro F1) on both the PubHealth and AVeriTec datasets, outperforming the \texttt{GPT-4} model. Furthermore, to address the high cost of explanation annotation, we generated 64 synthetic explanations from three large language models: \texttt{GPT-4-turbo}, \texttt{GPT-3.5-turbo}, and \texttt{Llama-3-8B} and few-shot fine-tune our step-1 model. The few-shot synthetic explanation fine-tuned model performed comparably to the fully fine-tuned self-rationalization model, demonstrating the potential of low-budget learning with synthetic data. Our label-adaptive self-rationalization approach presents a promising direction for future research on real-world explainable fact-checking with different labeling schemes\footnote{The source code and data can be found here: \url{https://github.com/jingyng/label-adaptive-self-rationalization}.}.
\end{abstract}


\section{Introduction}
Explainable fact verification is key for modern automated fact-checking. Recent fact-checking datasets usually contain annotated explanations~\cite{kotonya2020explainable,schlichtkrull2024averitec} to address its importance. However, research on explainable fact-checking methods mainly focuses on text summarization~\cite{atanasova2020generating,kotonya2020explainable,russo2023benchmarking} and, in such cases, explanations as summaries are not representative of real-world fact-checking explanations as they are not comparing the differences between claim and evidence to make conclusions. 
Self-rationalization, in turn, whereby models are trained to produce predictions and natural language explanations jointly, is a mainstream explainable approach for Natural Language Inference (NLI) tasks and could be used to further improve fact verification explanations~\cite{anonymous2024}. Notwithstanding, self-rationalization in its typical formulation is conditional to the target dataset labels being part of the language model pre/training~\cite{marasovic2022few,anonymous2024}.

\begin{figure}[t!]
    \centering
    \includegraphics[width=0.49\textwidth]{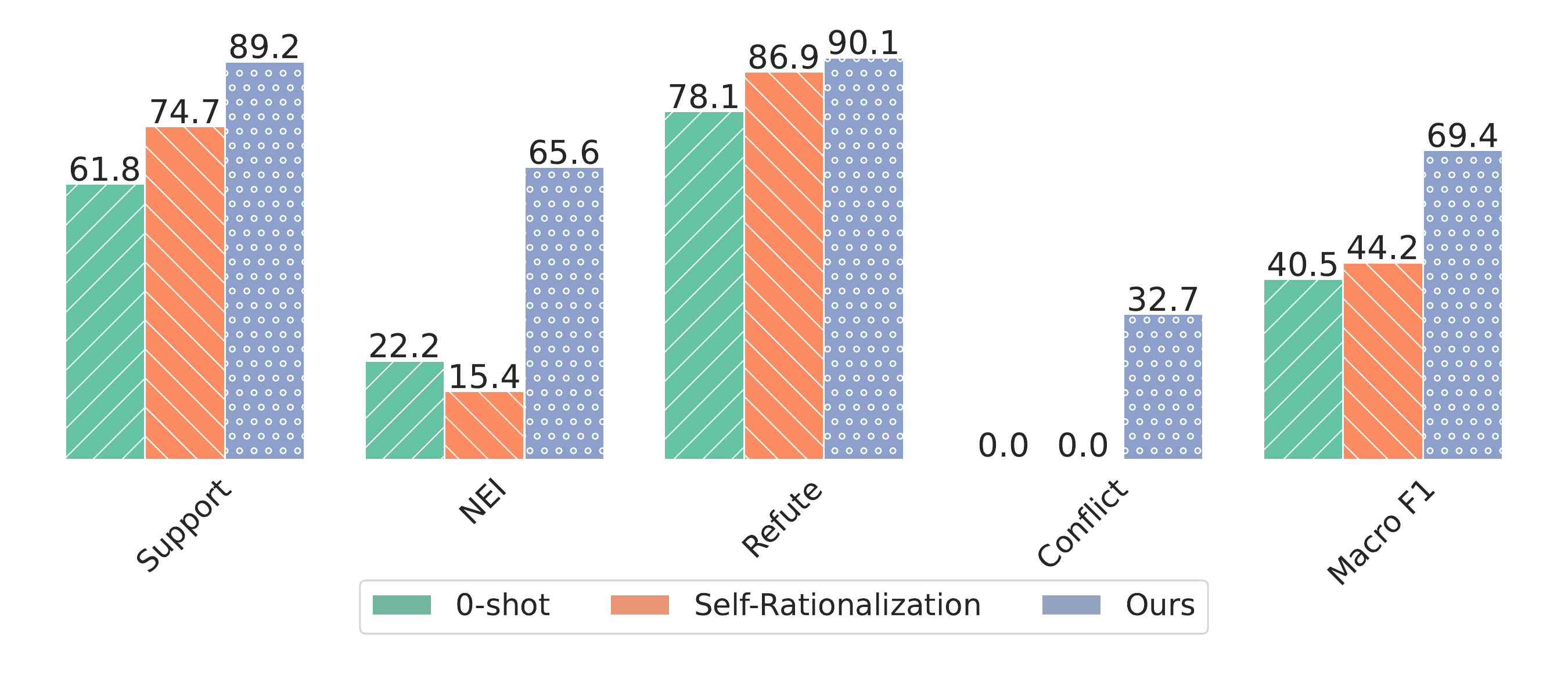}
    \caption{Models' performance on the AVeriTec dataset for each class (F1 score). 0-shot: zero-shot performance on \texttt{T5-3B};  Self-Rationalization: fine-tuned \texttt{T5-3B} model on joint labels and explanations. Ours: Label-adaptive Self-rationalization.}
    \label{fig:motivation}
\end{figure}

As an example, consider Figure~\ref{fig:motivation}. It depicts different methods performances on a recently released fact-checking dataset AVeriTec~\cite{schlichtkrull2024averitec}. This dataset comprises four labels, besides the typical 3-class label (Support, Not Enougn Info (NEI), Refute), it includes a new one for ``Conflict (Conflicting Evidence)''. When performing zero-shot on the \texttt{T5-3B} (green bars), a model pre-trained with NLI datasets (fact verification is often considered similar to NLI) shows reasonable results on the ``Support'' and ``Refute'' classes but performs poorly on ``NEI'', and completely fails on the new ``Conflict'' class. Self-rationalization fine-tuned on \texttt{T5-3B}, depicted by the orange bars in Figure~\ref{fig:motivation}, fails to learn the new class, resulting in low veracity prediction performance.

This problem is significant because most fact-checking datasets (e.g., FEVER~\cite{thorne2018fever}) usually label claim veracity with three classes: SUPPORT, REFUTE, and NEI (not enough information), which is comparable to NLI labels (entailment, contradiction, and neutral). However, many real-world fact-checking datasets usually have different labeling schemes with the number of classes varying from 2-27 classes~\cite{augenstein2019multifc} in some cases. As the labeling scheme shifts from NLI tasks, directly applying self-rationalization with models pre-trained on NLI datasets performs poorly for fact checking.

In this context, we propose a label-adaptive self-rationalization approach to tackle the challenge of the labeling shift for fact verification/checking. We first fine-tune a pre-trained model to learn the classification task with different labels; then, we fine-tune it again with labels and explanations to learn the self-rationalization task (explanations). Our results show that the 2-step formulation significantly outperforms direct self-rationalization learning by more than 20 percentage points (on the AVeriTec dataset) (Figure~\ref{fig:motivation}). This approach also achieves the best results compared to state-of-the-art methods. 

In summary, our contributions herein are twofold:
\begin{itemize}
    \item We propose a 2-step self-rationalization approach custom-tailored to the fact-checking domain;
    \item We propose to generate few-shot synthetic explanations by LLMs for step-2 self-rationalization, in case of lacking annotated explanations. In this case, the model's performance is comparable with the entire dataset. 
\end{itemize}

\section{Related Work}
This section presents available explainable fact-checking datasets in the literature and the most principled methods proposed to deal with this problem thus far. 

\subsection{Explainable Fact-checking Datasets}
Explainability has been an important research front in fact-checking; however, only a few datasets are constructed for this task. LIAR-PLUS~\cite{alhindi2018your} was the first dataset by extending the LIAR~\cite{wang2017liar} dataset with extracted justifications from PolitiFact fact-checking articles. Kotonya et al.~\cite{kotonya2020explainable} constructed a large dataset called PubHealth with claims about health topics collected from various fact-checking websites. e-FEVER~\cite{stammbach2020fever} was a dataset based on FEVER~\cite{thorne2018fever}, with synthetic explanations generated by \texttt{GPT-3}. A more recent dataset, AVeriTec, was released by Schlichtkrull et al.~\cite{schlichtkrull2024averitec}, in which the claims were also extracted from real-world fact-checking websites. Unlike previous explanation datasets, in which the explanations are summarized versions of the evidence, AVeriTec justifications are human-written explanations that reason over the retrieved evidence in the form of questions and answers.

\subsection{Explainable Fact-checking Methods}
 Summarization for explanation generation has been a popular approach, as most explanation datasets have their annotated explanations in the form of summarized evidence. Atanasova et al.~\cite{atanasova2020generating} first proposed an extractive summarization approach based on the LIAR-PLUS datasets. They generate fact-checking explanations by selecting important sentences from the original fact-checking ruling comments. Kotonya et al.~\cite{kotonya2020explainable} used a joint extractive-abstractive summarization approach to generate human-understandable explanations based on their PubHealth dataset. Russo et al.~\cite{russo2023benchmarking} benchmarked extractive and abstractive approaches and showed that performing an extractive approach before abstractive yielded the best result. The problem with the summarization approach is that the summaries cannot build clear connections between the claim and evidence to draw a conclusion. 

Another approach to generating explanations is through prompting of large language models (LLMs). Zhang et al.~\cite{zhang2023towards} proposed a prompting method (HiSS) to generate intermediate reasoning steps and a final prediction using \texttt{GPT-3.5}. The reasoning steps are composed of decomposed sub-claims followed by questions and answers related to each sub-claim. Zarharan et a.~\cite{zarharan2024tell} tested zero-/few shot abilities of LLMs on the PubHealth dataset, and they showed that parameter-efficient fine-tuning on the \texttt{Mixtral-7B} outperformed \texttt{GPT-4} model. The main issue with using LLMs is that their pre-training data are not transparent, which can cause data contamination, i.e., the test dataset might have been seen during their pre-training, causing unreliable performances. 

\section{Methodology}

\begin{figure*}[ht!]
    \centering
    \includegraphics[width=0.85\textwidth]{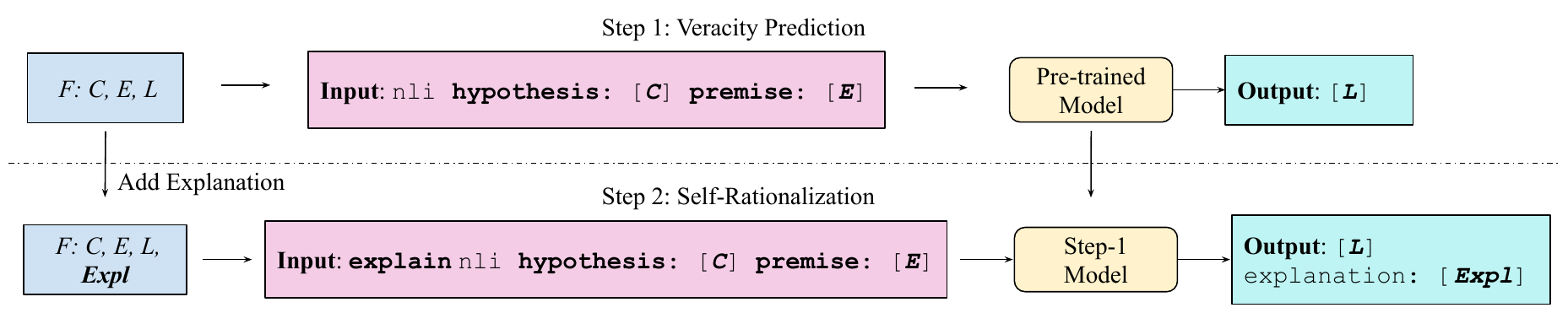}
    \caption{Label-adaptive self-rationalization 2-step pipeline. In step-1, the model learns veracity prediction with only provided labels; in Step-2, the model learns the self-rationalization task with both labels and explanations.}
    \label{fig:double-transfer-pinepine}
\end{figure*}

Provided with labels and explanations, directly fine-tuning for self-rationalization fails on newly added labels (as shown in Figure~\ref{fig:motivation}), thus we take a step-by-step approach to slowly adapt the model for the new domain and class. Our method is based on \texttt{T5-3B} model, as its size is comparable to many open large language models, and self-rationalization has been shown to perform well on T5 models~\cite{marasovic2022few,yordanov2022shot,anonymous2024}. 

\subsection{Label-Adaptive Self-rationalization Learning}

Our proposed approach is illustrated in Figure \ref{fig:double-transfer-pinepine}. It comprises two steps: in Step-1, the model learns to adapt to the new class with only provided labels; in Step-2, the model learns the self-rationalization task with labels and added explanations. We describe the details as follows:

Given a dataset $D=(C, E, L, Expl)$, with each sample $s_i=\{c_i, e_i, l_i, expl_i\}$, $c_i, e_i, l_i, expl_i$ represent a claim, evidence, label, and explanation, respectively, we perform two~steps.

\textbf{Step-1: Label Adaptation.} We first adapt and fine-tune the \texttt{T5} model to generate the veracity label $l_i$. Given the input $x_i=\{c_i, e_i\}$, we follow the same standard prompt template that was used to pre-train \texttt{T5} for the NLI task (``claims'' and ``evidence'' are mapped to ``hypothesis'' and ``premise'' as \texttt{T5} is more familiar with these words), as shown in first row of Figure~\ref{fig:double-transfer-pinepine}. 

\textbf{Step-2: Self-Rationalization.} After fine-tuning the model with the veracity prediction task, we now add gold explanation $expl_i$ to fine-tune the resulting \texttt{T5} model again after Step-1. Shown in second row of Figure~\ref{fig:double-transfer-pinepine}, we change the encoder prompt to add the word ``\textit{explain}'', and for the decoder prompt, a separation word ``\textit{explanation}'',  inspired~by~\cite{anonymous2024}. 

To simulate a realistic scenario with limited annotated explanations, we employ large language models (LLMs) to generate few-shot synthetic explanations. Specifically, we evaluate this task using \texttt{GPT-3.5-turbo-0125}, \texttt{GPT-4-turbo}, and \texttt{Llama-3-8B-Instruct}. We use the same prompt for generating the explanations with the three models, as shown~below:
    \begin{tcolorbox} [boxsep=0mm, left=1mm, right=1mm, top=1mm, bottom=1mm]
    \footnotesize
    \textit{\textbf{System}: You are a fact-checking assistant. You should not simply repeat the claim or evidence, your answer should be concise and short.} \\
    \textit{\textbf{User}: Given the evidence \{evidence\}, and claim \{claim\}. Please explain why the claim is \{ground truth label\}.}
    \end{tcolorbox}

\subsection{Data Processing and Label Mapping} 
We perform experiments on two datasets with explanation annotations: AVeriTeC~\cite{schlichtkrull2024averitec} and PubHealth~\cite{kotonya2020explainable}. We adopt these datasets as they better represent real-world fact-checking scenarios with 4-class annotations. 

\textbf{AVeriTeC:} The dataset comprises claims from 50 fact-checking organizations. It is unique in the way that the evidence in AVeriTeC is composed of questions and answers extracted from retrieval of online websites. To facilitate training, we concatenate the questions and answers as follows. Given a piece of evidence $e=\{q_1 (a_1,a_2,\cdots, a_i), \cdots, q_k (a_1,a_2,\cdots, a_j)\}$, we format it as ``$\text{Question 1}: q_1$ $ \text{Answer 1}: \{a_1 \text{ } a_2\cdots a_i\} \cdots $$\text{Question }k: q_k$ $\text{Answer }k: \{a_1 \text{ } a_2\cdots a_j\}$''. The justifications are human-annotated to reason over a claim's given questions and~answers.

\textbf{PubHealth:} The datasets contain claims from the health (biomedical) domain that are extracted from fact-checking and news review websites. The evidence consists of the full text from fact-checking articles or news reviews, with an average length exceeding 600 words, significantly longer than AVeriTeC’s average of 120 words. Explanations for claim veracity are provided through fact-checking justifications or news summaries.

We map the textual labels for different models as shown in Table~\ref{tab:labeling}. Specifically for \texttt{T5}, we align the labels with the NLI task naming scheme used during pre-training. For the ``\textit{Conflicting evidence}'' label in AVeriTeC, we equate it to the ``\textit{MIXTURE'}' class in PubHealth, which is ``\textit{partially true and false}'' for \texttt{GPT/Llama} models.

The data statistics for each dataset are shown in Table~\ref{tab:datasets}; we removed instances that contain empty claims. Both datasets have very imbalanced classes, with less data with ``\textit{NEI (not enough evidence)}'' and ``\textit{mixture}'' classes.

\begin{table}
\caption{Label mapping scheme} \label{tab:labeling}
    \renewcommand{\arraystretch}{1.2} 
    \large
    \begin{adjustbox}{width = 0.5\textwidth, center}
    \centering
    \begin{tabular}{ll|ll} 
        \toprule
        AVeriTeC &  PubHealth & \texttt{GPT/Llama} & \texttt{T5} \\ 
        \midrule
        Supported &  TRUE & true & entailment \\ \midrule
        \makecell[l]{Not enough \\evidence} &  UNPROVEN & \makecell[l]{not enough \\ information } & neutral \\ \midrule
        Refuted &  FALSE & false & contradiction \\ \midrule
        \makecell[l]{Conflicting evidence\\/cherry-picking} & MIXTURE & \makecell[l]{partially true \\ and false} & mixture \\ 
        \bottomrule
    \end{tabular} 
    \end{adjustbox}
\end{table}

\begin{table}[]
    \caption{Dataset details}
    \centering
    \begin{tabular}{lc|c} \toprule
          Dataset & \makecell{AVeriTec \\(Train / Dev)} & \makecell{PubHealth \\(Train / Dev / Test)} \\ \midrule
          entailment (S) &  848 / 122 & 5,078 / 629 / 599\\
          neutral (N) & 282 / 35 &291 / 41 / 45\\
          contradiction (R) & 1,742 / 305&3,001 / 380 / 388 \\
          mixture (M) & 195 / 38&1,434 / 164 / 201\\ \midrule
          \#words C & 17 / 17 &14 / 13 / 14 \\
          \#words E &113 / 122 & 714 / 708 / 718 \\\bottomrule
    \end{tabular}
        \label{tab:datasets}
\end{table}

\section{Experimental setup}

\subsection{Implementation Details}

In each fine-tuning experiment, we select the best model from the last epoch (50) without using a validation set. For AVeriTec, we use a batch size of 4 and a max input length of 512. For PubHealth, due to the length of the evidence, we use a batch size of 2 and a max input length of 1024. All experiments are based on NVIDIA A100 GPUs. For \texttt{GPT-4} zero-shot baseline, we set the temperature to be 0.7, with a max output length of 200.

\subsection{Evaluation Metrics}
To evaluate the veracity prediction and explanation quality, we first extract the label and explanation from the generated text using the separator ``\texttt{explanation: }''. For veracity prediction, we assess performance based on accuracy and macro F1 score. For explanations, we use both reference-based metrics (ROUGE scores and METEOR) and reference-free metrics. The latter is crucial in realistic scenarios where the test dataset lacks reference explanations for comparison. Specifically, we use the following reference-free metrics:

 \begin{itemize}
     \item Auto-J~\cite{li2023generative}: The metric is a model based on \texttt{LLaMA-2-13B-chat} by fine-tuning on judgments of LLM-generated responses with diverse user queries. It supports both single and pair-wise evaluations. We use it for single reference-free evaluations. The evaluation output comprises textual analysis and an overall quality rating between 1-10. 
     
     \item TigerScore~\cite{jiang2024tigerscore}: Another trained model-based metric that provides explainable evaluations for text generation tasks by following instructions. It outputs an overall error score ranging from 0 to -infinity, along with a textual analysis detailing the location and type of each detected error. We use the \texttt{TIGERScore-13B} model in our evaluation.
    
 \end{itemize}

For the reference-free metrics, the input must be formatted using instruction-based prompts. Our instructions are similar to those used for generating synthetic explanations with LLMs. We evaluate the explanations based on ground truth labels.

\subsection{Baselines}

We compare our two-step approach (denoted as \texttt{2-R}, with \texttt{R} denoting Rationalization) with the following baselines:

\begin{enumerate}

    \item \texttt{0-L}: zero-shot \texttt{T5-3B} baseline. As NLI datasets were used for \texttt{T5} pre-training, we formatted veracity prediction as an NLI task and prompted \texttt{T5-3B} to generate predictions. \texttt{L} denotes Label prediction.
    
    \item \texttt{1-R}: Compared to \texttt{2-R}, this baseline model is directly fine-tuned with labels and explanations without first fine-tuning for the veracity prediction task. 
    
    \item \texttt{1-L}: veracity prediction model fine-tuned with labels only (Step-1 model). The model cannot generate explanations, thus is not included for explanation comparison. 

    \item Baseline approach by Schlichtkrull~\cite{schlichtkrull2024averitec}. They have separate models for predicting veracity and generating explanations on the AVeriTec dataset, with the best results obtained with \texttt{BERT-Large} and \texttt{BART-Large}.

    \item Baseline approach by Kotonya~\cite{kotonya2020explainable}. They also have separate models for the two tasks; on the PubHealth dataset, the best results are based on \texttt{SCIBERT} and \texttt{BERT} models.

    \item Zarharan et al.~\cite{zarharan2024tell}: They studied different LLMs' performance on the PubHealth dataset. All their models are based on summarized evidence to reduce the evidence length, using 
    \texttt{GPT-3.5-turbo} for the summarization. The best results were achieved with parameter-efficient-fine-tuning (PEFT) on the \texttt{Mixtral-7B} model. 

    \item \texttt{GPT-4}. We conduct zero-shot prompting on \texttt{GPT-4-turbo} for the AVeriTec dataset. As reported in~\cite{zarharan2024tell}, \texttt{GPT-4}'s performance on the PubHealth dataset is directly reported in our work. We prompt the model to generate the output in JSON format to obtain the predicted veracity label and explanation, as illustrated~below.
\end{enumerate}

    \begin{tcolorbox}  [boxsep=0mm, left=1mm, right=1mm, top=1mm, bottom=1mm]
    \footnotesize
        \textit{\textbf{System}: You are a helpful assistant designed to output JSON, formatted as {``answer'':{}, ``reason:''{}}.} \\
        \textit{\textbf{User}: Based on the evidence, determine if the claim is true, false, not enough information to confirm, or partially true and false. \\
        Evidence: {[evidence]} \\
        Claim: {[claim]} \\
        Options: 
        - true 
        - not enough information 
        - false 
        - partially true and false \\
        Please provide your reason.}
    \end{tcolorbox}
    We directly refer to the numbers reported in the respective paper for baseline results. For explanation evaluation, Zarharan et al. ~\cite{zarharan2024tell} made their results publicly available, so we ran all evaluation metrics based on their released explanations.

\section{Results and Discussions}

We present results on veracity prediction and explanation generation in comparison with baselines; and the results of fine-tuning on few-shot synthetic LLM-explanations. 

\subsection{Veracity Prediction Performance}

 Table \ref{tab:accuracy} shows the veracity prediction results on different baseline models and our \texttt{2-R} model. As expected, \texttt{0-L} (zero-shot on \texttt{T5-3B}) cannot predict the class ``\textit{mixture}'' for either dataset. For AVeriTeC, our \texttt{2-R} model is comparable with \texttt{GPT-4}, with the best accuracy of 85.2\%, while being a much smaller model. For PubHealth, the \texttt{1-L} model achieved the best performance, while \texttt{2-R} model slightly dropped (2\%) on Macro F1 after learning to generate explanations. Both outperform the larger baseline models (\texttt{Mixtral-7B} and \texttt{GPT-4}). For both datasets, the \texttt{2-R} model improved performance (Macro F1) by more than 10 percentage points compared with the \texttt{1-R} model, showing that letting models learn the veracity task first greatly helps the model to adapt to the new domain with new classes. Specifically, the \texttt{1-R} model struggled with predicting classes ``\textit{neutral}'' and ``\textit{mixture}'', but with our label-adaptive approach (\texttt{2-R}), the model was able to improve predictions on these classes significantly. 

\begin{table}[h!]
\footnotesize
\centering
\caption{Performance (per class and overall) comparison on veracity prediction. }
\label{tab:accuracy}
\renewcommand{\arraystretch}{1.2}
\begin{adjustbox}{width = 0.5\textwidth, center}
\begin{tabular}{clcccc|cc} \toprule
 & Model & S & N & R & M & F1 & Acc. \\ \midrule
\multirow{5}{*}{\rotatebox{90}{AVeriTeC}}
& \texttt{BERT-Large}\cite{schlichtkrull2024averitec} & $48.0$ & $59.0$ & $74.0$ & $15.0$ & $49.0$ & $49.0$\\
& \texttt{GPT-4} & $\underline{83.5}$ & $\textbf{65.9}$ & $\textbf{91.5}$ & $\textbf{45.5}$ & $\textbf{71.6}$  & $83.0$ \\ \cline{2-8}
& \texttt{0-L} & $64.8$ & $22.2$ & $78.1$ & $0.0$ & $40.5$  & $62.0$\\
& \texttt{1-R} & $74.7$ & $15.4$ & $86.9$ & $0.0$ & $44.2$  & $76.2$\\
& \texttt{1-L} & $87.5$ & $59.0$ & $89.3$ & $29.5$ & $66.3$  & $\underline{83.4}$\\
& \texttt{2-R} & $\textbf{89.2}$ & $\underline{65.6}$ & $\underline{90.1}$ & $\underline{32.7}$ & $\underline{69.4}$  & $\textbf{85.2}$\\ \midrule
\multirow{5}{*}{ \rotatebox{90}{PubHealth} }
& \texttt{SCIBERT}\cite{kotonya2020explainable} & - & - & - & - & $70.5$  & $69.7$\\
& \texttt{Mistral-7B}\cite{zarharan2024tell} & $92.7$ & $48.6$ & $82.1$ & $\underline{57.1}$ & $70.1$  & $81.8$ \\
& \texttt{GPT-4}\cite{zarharan2024tell}  & $80.6$ & $18.2$ & $73.0$ & $42.0$ & $53.4$   & $69.6$\\ \cline{2-8}
& \texttt{0-L} & $65.0$ & $2.8$ & $42.9$ & $0.0$ & $27.7$  & $48.7$\\
& \texttt{1-R} & $91.2$ & $26.9$ & $79.8$ & $38.9$ & $59.2$  & $79.1$\\
& \texttt{1-L}& $\textbf{93.4}$ & $\textbf{60.5}$ & $\textbf{84.2}$ & $\textbf{58.2}$ & $\textbf{74.1}$  & $\textbf{83.7}$\\ 
& \texttt{2-R}& $\underline{93.2}$ & $\underline{57.5}$ & $\underline{83.4}$ & $55.1$ & $\underline{72.3}$  & $\underline{83.1}$\\  \bottomrule
\end{tabular}
\end{adjustbox}
\end{table}

\subsection{Generated Explanation Quality}

We show the evaluation of generated explanation quality in Table~\ref{tab:performance_auto}. For both datasets, \texttt{GPT-4} generated explanations have the best scores on the reference-free metrics, indicating the reasoning abilities of \texttt{GPT-4}, although it has a tendency to be verbose (having the longest explanations on average). Our \texttt{2-R} approach has the highest ROUGE scores, outperforming the baselines. For the AVeriTec dataset, the \texttt{2-R} model generates better explanations than the \texttt{1-R} model, as agreed by all metrics. For the PubHealth dataset, the scores for the two models are very similar, and both have the highest ROUGEs and METEOR scores. In general, the results show that fine-tuned models generate explanations that are better aligned with reference explanations, as the training data follow a similar~pattern. 

\begin{table}[ht]
\centering
\caption{Explanation evaluation with reference-free and reference-based metrics. \#W means the average number of words in the explanations. Reference means gold explanation.}
\label{tab:performance_auto}
\renewcommand{\arraystretch}{1.2} 
\begin{adjustbox}{width = 0.5\textwidth, center}
\begin{tabular}{cl|cc|cc|c} \toprule
& Model & AJ$\uparrow$ & Tiger$\downarrow$ & ROUGEs & METEOR & \#W \\ \midrule
\multirow{5}{*}{ \rotatebox{90}{AVeriTeC}} 
& \texttt{BART-Large}\cite{schlichtkrull2024averitec} & - & - & - & \textbf{.28} & - \\
& \texttt{GPT-4} & \textbf{4.99} & \textbf{0.64} & 25 / 9 / 19 & .31 & 60 \\ \cline{2-7}
& \texttt{1-R} & 3.45 & 2.06 & 27 / 10 /23 & .24 & 18 \\ 
& \texttt{2-R (Ours)} & 3.61 & 1.87 & \textbf{29 / 12 / 25} & .26 & 18 \\  \cline{2-7} 
& Reference & 3.54 & 1.48 & - & - & 22 \\ \midrule
\multirow{6}{*}{ \rotatebox{90}{PubHealth} }
& \texttt{BERT}\cite{kotonya2020explainable} & -  & - & 32 / 13 / 27 & - & - \\
& \texttt{Mistral-7B}\cite{zarharan2024tell} & 3.99 & 1.88 & 36 / 15 / 26 & .29 & 73 \\ 
& \texttt{GPT-4}\cite{zarharan2024tell} & \textbf{4.80} & \textbf{0.53} & 26 / 8 / 17 & .24 & 75 \\ \cline{2-7}
& \texttt{1-R} & 3.63 & 2.34 & 43 / 24 / 34 & \textbf{.37} & 59 \\
& \texttt{2-R (Ours)} & 3.62 & 2.50 & \textbf{43 / 24 / 35} & \textbf{.37} & 59 \\ \cline{2-7} 
& Reference & 3.70 & 1.23 & - & - & 76 \\ \bottomrule
\end{tabular} 
\end{adjustbox}
\end{table}

Overall, our \texttt{2-R} approach achieves the highest veracity prediction performance and the best reference-based scores for explanations, outperforming LLMs and other state-of-the-art baselines.

\subsection{Results from Synthetic Few-shot Explanations}

To demonstrate the potential of our two-step approach in data-scarce scenarios, we test \texttt{Step-2} with few-shot fine-tuning. We select 16 samples per class (64 samples total) to prompt an LLM to generate synthetic explanations. These samples and their generated explanations are then used to fine-tune the \texttt{1-L} model. For robust results, we select few-shot samples with three different random seeds and report the results in average and standard deviation. The results for veracity prediction and explanation generation are shown in Table \ref{tab:syn_label} and \ref{tab:syn_expl}.

The veracity prediction results show that \texttt{Step-2} with very few amount of data still achieve much better performance than end-to-end self-rationalization model (\texttt{1-R}), and perform comparably to the \texttt{2-R} with full dataset fine-tuning. In terms of explanation quality, the reference-free metrics indicate that the best explanations are from the \texttt{2-R(GPT-3.5)}, with a similar Auto-J score compared to the best, and the lowest TigerScore among few-shot models. 

Surprisingly, the \texttt{2-R(GPT-4)} model performs worse than both \texttt{2-R(GPT-3.5)} and \texttt{2-R(Llama-3-8B)}, in contrast to Table~\ref{tab:performance_auto}, where \texttt{GPT-4} model generated explanations are much better. We hypothesize that when generated text is long (\texttt{2-R(GPT-4)} model explanations are almost twice as long compared with the rest), it is more detailed but also more likely to contain errors. 

We show an example of explanations generated by different models from the PubHealth dataset in Figure~\ref{tab:exam_pubhealth}). We see that as the explanation becomes longer, models tend to hallucinate and makes more errors. In this sense, \texttt{GPT-3.5} and \texttt{Llama-3-8B} generated explanations are better for having shorter explanations and thus less likely to make errors. This gap is particularly captured by TigerScore (Table~\ref{tab:performance_auto}), which measures the number of errors in the explanations. 

\begin{table}[ht]
\centering
\caption{Veracity prediction results with few-shot \texttt{Step-2} fine-tuning under different LLM-based synthetic explanations. All models are based on \texttt{T5-3B}. Orig. means original annotated explanations. Full means entire dataset fine-tuning, otherwise few-shot fine-tuning.}
\label{tab:syn_label}
\large
\begin{adjustbox}{width = 0.5\textwidth, center}
\renewcommand{\arraystretch}{1.2} 
\begin{tabular}{clcccc|c} \toprule
& Expl. Source & S & N & R & M & F1 \\ \midrule
\multirow{5}{*}{ \rotatebox{90}{AVeriTeC} } & \texttt{2-R(GPT-4)} & $83.1_{\pm{2.0}}$ & $52.9_{\pm{3.9}}$ & $86.4_{\pm{0.7}}$ & $29.8_{\pm{4.3}}$ & $63.1_{\pm{1.1}}$ \\
& \texttt{2-R(GPT-3.5)} & $86.2_{\pm{2.3}}$ & $\textbf{61.0}_{\pm{1.6}}$ & $85.3_{\pm{2.2}}$ & $\textbf{35.1}_{\pm{3.3}}$ & $\textbf{66.9}_{\pm{1.0}}$ \\
& \texttt{2-R(Llama-3-8B)} & $83.0_{\pm{5.2}}$ & $58.1_{\pm{5.7}}$ & $85.9_{\pm{0.4}}$ & $35.0_{\pm{4.2}}$ & $65.5_{\pm{2.1}}$ \\
& \texttt{2-R} (orig.) & $\textbf{86.5}_{\pm{2.8}}$ & $58.3_{\pm{4.1}}$ & $\textbf{87.2}_{\pm{0.7}}$ & $30.6_{\pm{2.6}}$ & $65.6_{\pm{1.0}}$ \\
& \texttt{1-R} (orig., Full) & $74.7$ & $15.4$ & $86.9$ & $0.0$ & $44.2$ \\
& \texttt{2-R} (orig., Full) & $89.2$ & $65.6$ & $90.1$ & $32.7$ & $69.4$ \\ \midrule
\multirow{5}{*}{ \rotatebox{90}{PubHealth} } & \texttt{2-R(GPT-4)} & $86.9_{\pm{1.0}}$ & $38.6_{\pm{2.0}}$ & $75.8_{\pm{1.6}}$ & $54.4_{\pm{1.1}}$ & $63.9_{\pm{1.2}}$ \\
& \texttt{2-R(GPT-3.5)} & $87.5_{\pm{1.5}}$ & $42.9_{\pm{2.3}}$ & $76.3_{\pm{1.6}}$ & $\textbf{55.5}_{\pm{1.6}}$ & $65.5_{\pm{1.0}}$ \\
& \texttt{2-R(Llama-3-8B)} & $\textbf{88.9}_{\pm{1.1}}$ & $46.1_{\pm{3.0}}$ & $\textbf{78.6}_{\pm{2.6}}$ & $54.3_{\pm{2.2}}$ & $\textbf{67.0}_{\pm{1.4}}$ \\
& \texttt{2-R} (orig.) & $86.1_{\pm{0.5}}$ & $\textbf{46.7}_{\pm{2.4}}$ & $78.1_{\pm{2.3}}$ & $51.9_{\pm{0.3}}$ & $65.7_{\pm{1.1}}$ \\
& \texttt{1-R} (orig., Full)& $91.2$ & $26.9$ & $79.8$ & $38.9$ & $59.2$ \\
& \texttt{2-R} (orig., Full)  & $93.2$ & $57.5$ & $83.4$ & $55.1$ & $72.3$ \\ \bottomrule
\end{tabular}
\end{adjustbox}
\end{table}

\begin{table}[ht]
\centering
\caption{Explanation evaluation results with \texttt{Step-2} few-shot fine-tuning under different LLM-based synthetic explanations. All models are based on \texttt{T5-3B}.}
\label{tab:syn_expl}
\Large
\begin{adjustbox}{width = 0.5\textwidth, center}
\renewcommand{\arraystretch}{1.2} 
\begin{tabular}{clcc|cc|c} \toprule
& Expl. Source & Auto-J$\uparrow$ & Tiger$\downarrow$ & ROGUE-1 / 2 / L & METEOR & \#W. \\ \midrule
\multirow{6}{*}{ \rotatebox{90}{AVeriTeC}  } & \texttt{2-R(GPT-4)} & $\textbf{4.51}_{\pm{0.04}}$ & $4.35_{\pm{0.45}}$ & $21_{\pm{0.7}}$ / $8_{\pm{0.3}}$ / $16_{\pm{0.4}}$ & $.29_{\pm{0.0}}$ & $84_{\pm{5}}$ \\
& \texttt{2-R(GPT-3.5)} & $4.42_{\pm{0.10}}$ & $\textbf{2.49}_{\pm{0.25}}$ & $\textbf{27}_{\pm{0.2}}$ / $10_{\pm{0.0}}$ / $\textbf{20}_{\pm{0.1}}$ & $\textbf{.30}_{\pm{0.0}}$ & $45_{\pm{1}}$ \\
& \texttt{2-R(Llama-3-8B)} & $4.31_{\pm{0.13}}$ & $2.90_{\pm{0.65}}$ & $27_{\pm{1.2}}$ / $\textbf{11}_{\pm{0.6}}$ / $20_{\pm{0.9}}$ & $.29_{\pm{0.0}}$ & $45_{\pm{4}}$ \\
& \texttt{2-R} (orig.) & $3.38_{\pm{0.06}}$ & $2.70_{\pm{0.15}}$ & $25_{\pm{0.7}}$ / $8_{\pm{0.4}}$ / $20_{\pm{0.5}}$ & $.22_{\pm{0.0}}$ & $22_{\pm{3}}$ \\
& \texttt{1-R} (orig., Full)& $3.45$ & $2.06$ & $27$ / $10$ /$23$ & $.24$ & $18$ \\ 
& \texttt{2-R} (orig., Full) & $3.61$ & $1.87$ & $29$ / $12$ / $25$ & $.26$ & $18$ \\ \cline{2-7}
& Reference & $3.54$ & $1.48$ & - & - & $22$ \\ \midrule
\multirow{6}{*}{ \rotatebox{90}{PubHealth}  } & \texttt{2-R(GPT-4)} & $\textbf{4.47}_{\pm{0.03}}$ & $4.27_{\pm{0.30}}$ & $24_{\pm{0.3}}$ / $7_{\pm{0.0}}$ / $16_{\pm{0.1}}$ & $\textbf{.25}_{\pm{0.0}}$ & $119_{\pm{3}}$ \\
& \texttt{2-R(GPT-3.5)} & $4.39_{\pm{0.01}}$ & $\textbf{2.34}_{\pm{0.23}}$ & $27_{\pm{0.3}}$ / $9_{\pm{0.2}}$ / $18_{\pm{0.2}}$ & $.24_{\pm{0.0}}$ & $68_{\pm{5}}$ \\
& \texttt{2-R(Llama-3-8B)} & $4.29_{\pm{0.03}}$ & $2.89_{\pm{0.11}}$ & $27_{\pm{0.5}}$ / $8_{\pm{0.2}}$ / $18_{\pm{0.3}}$ & $.23_{\pm{0.0}}$ & $58_{\pm{2}}$ \\
& \texttt{2-R} (orig.) & $3.65_{\pm{0.03}}$ & $2.84_{\pm{0.26}}$ & $\textbf{31}_{\pm{0.6}}$ / $\textbf{12}_{\pm{1.0}}$ / $\textbf{22}_{\pm{0.9}}$ & $\textbf{.25}_{\pm{0.0}}$ & $56_{\pm{5}}$ \\
& \texttt{1-R} (orig., Full) & $3.63$ & $2.34$ & $43$ / $24$ / $34$ & $.37$ & $59$ \\
& \texttt{2-R} (orig., Full) & $3.62$ & $2.50$ & $43$ / $24$ / $35$ & $.37$ & $59$ \\ \cline{2-7}
& Reference & $3.70$ & $1.23$ & - & - & $76$ \\ \bottomrule
\end{tabular}
\end{adjustbox}
\end{table}

\begin{table}[ht!]
\caption{An example of generated explanations from different models on PubHealth dataset. In the evidence and explanations, the accurate and relevant text is highlighted in color \textcolor{teal}{teal}, while the inaccurate or hallucinated text is highlighted in \textcolor{red}{red}. Best viewed in color.} \label{tab:exam_pubhealth}
\centering
\begin{adjustbox}{width = 0.5\textwidth, center}
\Large
\begin{tabular}{@{}p{\textwidth}@{}}
\hline
\textbf{Claim} (Mixture) \\ The Westmoreland News ran a KKK ad on the front page of their newspaper in April 2018. \\\hline
\textbf{Evidence} (unimportant parts are omitted due to its length) \\
On 25 April 2018, the front page story of Virginia’s Westmoreland News described the disturbing occurrence of KKK recruitment flyers in front home homes in Colonial Beach, VA [...] \textcolor{teal}{In a move that has generated significant online controversy, the Westmoreland News reproduced a full, un-redacted version of that ad in their story, leading some to brand it has free advertising for the KKK}. As reported by Newsweek and local CBS affiliate WTVR, many readers of the paper were outraged: They posted the whole flyer as if they were giving them free advertising,” Colonial Beach resident Betty Tate Thompson told WTVR. “It even had the number on flyer to contact the head of the KKK. Which I thought was totally ridiculous.” [...] Plott said the editor called him back with an explanation. “She explained that she wanted to put it in to show people that those of us in Colonial Beach who may think there’s not racism; there is and I think she didn’t expect for this to blow up the way it did and with this outcome,” explained Plott. \textcolor{teal}{While the ad was reproduced in full, The Westmoreland News did post a disclaimer on the story with this text: Westmoreland News in no way condones or supports the content or message of this flyer, nor does it condone or support any branch of the Loyal White Knights, or the KKK}. Decisions on how to report the occurrence of KKK flyers in the media have not been made consistently in the past. [...] In each of those cases, it should be noted, the flyer was reproduced as an obvious photograph of a paper document. \textcolor{teal}{The front page image used by the Westmoreland Times, however, could be confused for a paid advertisement. Because that was not the case here, however, we rank the claim that the newspaper ``ran an ad'' for the KKK as a mixture}.  \\\hline
\textbf{\texttt{GPT-4}}                                   \\
The context refers to an occurrence in April 2018 where the Westmoreland News reproduced a KKK recruitment ad on its front page. The controversy discussed was \textcolor{red}{about the decision to run this ad}, which was perceived as free advertising for the KKK. Therefore, \textcolor{red}{the claim is \textbf{supported} by the context}.                            \\ \hline
\textbf{\texttt{1-R} (orig., Full) }                     \\
\textcolor{teal}{What's true: The Westmoreland News reproduced a full, un-redacted version of a KKK recruitment flyer in their front page story}. \textcolor{teal}{What's false}: The newspaper \textcolor{red}{did not pay for the~ad}.                                  \\ \hline
\textbf{\texttt{2-R} (orig., Full) }                                 \\
\textcolor{teal}{What's true: The Westmoreland News reproduced a full, un-redacted version of a KKK recruitment flyer in their front page story. What's false: The newspaper did not run an ad for the KKK}.                                                                \\ \hline
\textbf{\texttt{2-R(GPT-4)}}                                 \\
The claim that the Westmoreland News ran a KKK ad on the front page of their newspaper in April 2018 is \textcolor{teal}{\textbf{mixture} because the evidence provided does not conclusively establish that the newspaper provided free advertising for the KKK. While the publication did reproduce a full, un-redacted version of the KKK ad in their story, the inclusion of the contact information and phone numbers on the flyer raised concerns among readers and sparked outrage on social media}. Additionally, \textcolor{red}{the editorial decision to run the ad in its entirety was made \textbf{after} significant discussion and debate amongst community members and law enforcement}, who objected to the inclusion of such explicit recruitment materials in a news story. Therefore, \textcolor{red}{the claim cannot be fully confirmed as it is made with certainty.} \\ \hline
\textbf{\texttt{2-R(GPT-3.5)}}                               \\
The claim that The Westmoreland News ran a KKK ad on the front page of their newspaper in April 2018 is \textcolor{teal}{\textbf{partially true} because the newspaper included a full, un-redacted version of the ad in their report on the occurrence of KKK recruitment flyers} in Colonial Beach, VA. \textcolor{red}{The claim is \textbf{false}} because \textcolor{teal}{the newspaper did not explicitly state that it was running an ad for the KKK}, and \textcolor{red}{the claim is also \textbf{false}} because \textcolor{teal}{the editor provided a disclaimer on the article}.                 \\ \hline
\textbf{\texttt{2-R(Llama-3-8B)}}                               \\
The claim is \textcolor{teal}{\textbf{partially true} because the Westmoreland News did reproduce a full, un-redacted version of the KKK ad in their story}, which generated significant online controversy. \textcolor{teal}{While the newspaper did not explicitly endorse the KKK, it did state that it did not condone or support the group's message or message of racism. Additionally, the claim is \textbf{partially false}} because the newspaper \textcolor{red}{did not provide sufficient evidence to prove the claim}.                \\ \hline
\textbf{Reference (Gold)}                               \\
What's true: The Westmoreland News did publish a copy of a KKK flyer in full on their front page as part of their reporting on KKK flyers being distributed in the region. What's false: The document was not run as an ad in the paper, but was intended to be an illustration for their coverage of the KKK.                 \\ \hline
\end{tabular}%
\end{adjustbox}
\end{table}

\section{Conclusions, Limitations and future work}
We proposed an effective two-step approach for joint fact-verification and explanation generation with self-rationalization. Our results show that having a label prediction step significantly helped the model to adapt to new classes and perform better. Our method with \texttt{T5-3B} outperformed larger models, including \texttt{Mixtral-7B} and \texttt{GPT-4}. We further utilized LLMs to generate few-shot synthetic explanations to fine-tune our \texttt{T5-3B} model, and it outperformed end-to-end self-rationalization models fine-tuned on the entire dataset. We also show that \texttt{T5-3B} models struggle with generating longer explanations when learning from \texttt{GPT-4} explanations. 

\paragraph{Limitations}{Our work nonetheless has its limitations. 1) When using a generative model for classification, the naming of labels is factor that affects performance, as different models may have their own way of formatting the labels during pretraining. 2) We use the same instructions for different LLM models, but there may be other instructions that help them generate more accurate explanations. 3) Our method is based on the encoder-decoder architecture, so it may not be generalized to decoder-only architecture models. 4) Our explanation evaluation is based on automatic metrics, although they have been shown to correlate well with humans, they were not specifically designed to evaluate generated explanations for fact verification.}

\paragraph{Future work}{Future work may focus on studying what models/instructions can generate better synthetic explanations for smaller models to learn from. Moreover, testing the approach to multilingual models and datasets is also a promising endeavor.}

\bibliographystyle{IEEEtran}
\bibliography{IEEEabrv,main}

\end{document}